\documentclass[10pt,twocolumn,letterpaper]{article}
\usepackage[T1]{fontenc}

\usepackage[pagenumbers]{wacv} 

\usepackage{multirow}
\usepackage{colortbl}
\usepackage{pifont}

\definecolor{lightblue}{rgb}{0.796,0.894,0.9808}
\definecolor{forestgreen}{RGB}{34,139,34}
\definecolor{crimson}{RGB}{220,20,60}

\definecolor{wacvblue}{rgb}{0.21,0.49,0.74}
\usepackage[pagebackref,breaklinks,colorlinks,hyperfootnotes=false,allcolors=wacvblue]{hyperref}

\def\wacvPaperID{3037}
\def\confName{WACV}
\def\confYear{2027}
\hypersetup{pdftitle={PTQ4SAM3: Semantics-Preserving Post-Training Quantization for Segment Anything Model 3},
pdfauthor={Xi Xiao, Yunbei Zhang, Lin Zhao, Janet Wang, Xinru Jiang, Yanshu Li, Chenrui Ma, Pan Wang, Tianyang Wang, Xiao Wang}}

\title{PTQ4SAM3: Semantics-Preserving Post-Training Quantization for Segment Anything Model 3}

\author{%
Xi Xiao$^{1}$\quad Yunbei Zhang$^{2}$\quad Lin Zhao$^{3}$\quad Janet Wang$^{2}$\quad Xinru Jiang$^{3}$\\
Yanshu Li$^{4}$\quad Chenrui Ma$^{5}$\quad Pan Wang$^{6}$\quad Tianyang Wang$^{1}$\quad Xiao Wang$^{7}$\\
$^{1}$University of Alabama at Birmingham, USA\\
$^{2}$Tulane University, USA\quad $^{3}$Northeastern University, USA\\
$^{4}$University of Texas at Austin, USA\quad $^{5}$University of Virginia, USA\\
$^{6}$University of Pittsburgh, USA\quad $^{7}$Oak Ridge National Laboratory, USA\\\texttt{wangx2@ornl.gov}\quad\texttt{tw2@uab.edu}
}

\begin{document}
\twocolumn[{
\maketitle
\vspace{-1.4em}
\begin{center}
    \includegraphics[width=0.85\textwidth]{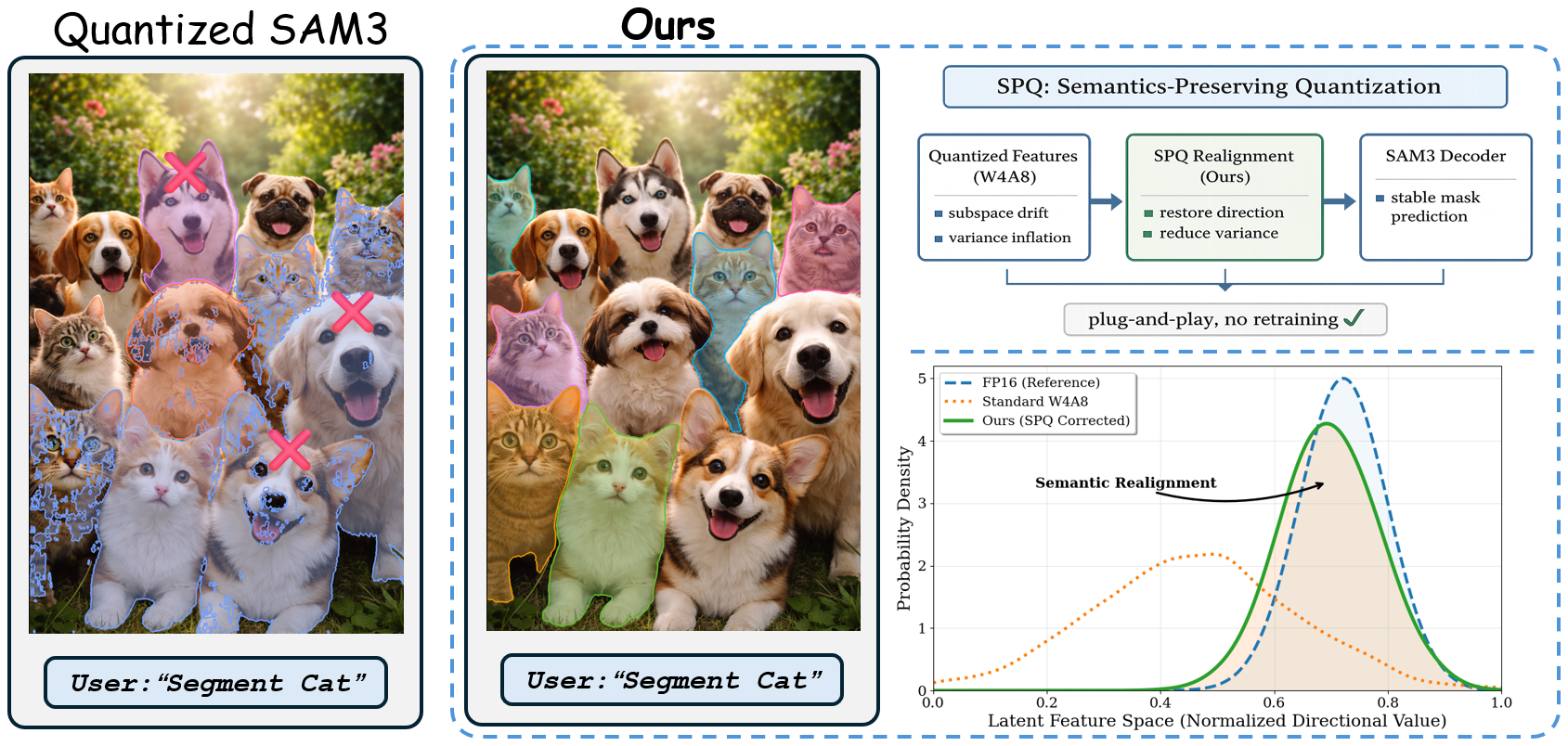}
    \captionof{figure}{\textbf{SPQ preserves prompt grounding after quantization.} Under the same text prompt, standard W4A8 quantization can select the wrong objects (left, red crosses). SPQ calibrates the model with text concepts, preserves residual features, and corrects memory bias (right). The latent-feature diagram illustrates features moving closer to the FP16 reference during offline calibration.}
    \label{fig:teaser}
\end{center}
\vspace{0.4em}
}]

\begingroup
\renewcommand{\thefootnote}{}
\footnotetext{\copyright\ 2027 IEEE. Personal use of this material is permitted. Permission from IEEE must be obtained for all other uses, in any current or future media, including reprinting/republishing this material for advertising or promotional purposes, creating new collective works, for resale or redistribution to servers or lists, or reuse of any copyrighted component of this work in other works.}
\endgroup

\begin{abstract}
Segment Anything Model~3 (SAM~3) can segment objects from text prompts and track them across video, but its memory and computation costs make deployment difficult. Post-training quantization (PTQ) reduces these costs by using fewer bits. However, a quantized model can select the wrong object or lose track of its target across frames. We call this prompt-alignment drift. Controlled tests show that perturbing low-variance feature directions strongly affects target selection. We propose Semantics-Preserving Quantization (SPQ), which gives more calibration weight to concept-sensitive parameters, preserves residual features, and corrects the mean error in video memory. Fixed text concepts guide calibration without task annotations. All three components operate offline and add no inference operators. With a W4A8 image encoder, SPQ improves RefCOCO Pr@0.5 from 42.1 to 72.5 over PTQ4SAM while retaining 54.8 COCO mask AP. Experiments on LVIS, RefCOCO variants, and SA-V show consistent gains. A CLIP study also shows improved image--text retrieval.

\end{abstract}

\section{Introduction}
\label{sec:intro}

Segment Anything Model (SAM)~\cite{kirillov2023segment} lets users select objects through simple prompts. SAM~3~\cite{carion2025sam} extends this ability to text prompts and video: a user can name an object, obtain its mask, and track it across frames. These capabilities make the model useful for a wide range of visual tasks. They also make deployment expensive. The model must store large image features, match them with text, and maintain a memory of past frames. Reducing these costs is part of a broader challenge in efficient multimodal learning~\cite{wang2026models}.

Post-training quantization (PTQ) offers a practical way to reduce model size and inference cost. It converts a pretrained model to lower precision using a small calibration set~\cite{nagel2020up,nagel2021white}. Existing methods improve low-bit vision transformers and SAM through better reconstruction, rounding, and calibration~\cite{liu2021post,linfq,li2023repq,zhong2024erq,ma2024outlier,moon2024instance,shang2024quantized,xulogart,lv2024ptq4sam,zhang2026saqsam}. Related methods also support large language models~\cite{li2022q,xiao2023smoothquant,frantar2023optq,MLSYS2024_42a452cb}. For SAM~3, however, preserving mask quality is only part of the problem. The quantized model must also preserve the connection between a prompt and its intended object.

We find that a quantized SAM~3 can draw a plausible mask around the wrong object, omit a prompted part, or switch identities as a video unfolds. These errors occur even when feature reconstruction and standard segmentation scores remain strong. We call this behavior \emph{prompt-alignment drift}. Fig.~\ref{fig:teaser} shows how the same text prompt can produce different target selections after quantization. This observation motivates a central question: which feature errors matter most to prompt grounding?

We study this question through the dominant principal directions of model features and their low-variance residual complement. Matched perturbations in residual directions reduce grounding more, while perturbations in dominant directions reduce mask AP more. Feature removal gives a similar task contrast, and the random-basis control has intermediate effects (Sec.~\ref{subsec:mechanism_checks} and Supplementary Sec.~B). We use a subspace perturbation view~\cite{davis1970rotation,yu2015useful} to organize this analysis and guide calibration toward the residual features that grounding depends on.

Based on these observations, we propose \textbf{Semantics-Preserving Quantization (SPQ)}. SPQ combines three calibration steps. \textbf{Semantic-Anchor Calibration} gives more weight to parameters that affect text-concept scores. \textbf{Orthogonal Subspace Rectification} preserves features along the residual directions. \textbf{Temporal Distribution Alignment} estimates the mean error in video memory and folds its correction into existing biases. These components build on second-order PTQ weighting and static distribution correction~\cite{dong2019hawq,wu2025fima,xiao2023smoothquant}. Together, they preserve prompt grounding and temporal memory while keeping the inference graph unchanged.

We evaluate SPQ on image segmentation, referring segmentation, and video tracking, and test its transfer to CLIP. With a W4A8 image encoder, SPQ reaches 54.8 COCO mask AP and 72.5 RefCOCO Pr@0.5, compared with 54.1 and 42.1 for PTQ4SAM. Controlled baseline comparisons and a complete factorial ablation show how the three components contribute across tasks. Our contributions are threefold:
\begin{itemize}
    \item We identify prompt-alignment drift in quantized SAM~3 and show that grounding is especially sensitive to perturbations in low-variance residual directions.
    \item We introduce SPQ, which combines concept-sensitive calibration, residual-feature matching, and an offline correction for video memory.
    \item We evaluate these components across image, referring, and video tasks, with controlled comparisons and a CLIP transfer study.
\end{itemize}

\section{Preliminaries}
\label{sec:preliminaries}

We first describe standard PTQ, then examine how feature errors affect prompt grounding. We use principal components to separate dominant feature directions from their low-variance residual complement and measure their sensitivity through controlled interventions (Supplementary Sec.~B). MedSTP preserves the dominant visual-token subspace during medical-VQA pruning~\cite{fan2027medstp}. SPQ focuses on quantization error in the residual feature directions of SAM~3.

\subsection{Revisiting Standard PTQ}
\label{subsec:revisit_ptq}

Post-training quantization (PTQ)~\cite{nagel2020up,librecq} converts a pretrained FP16 model into a low-bit model using a small calibration set $\mathcal{D}_{\text{calib}}$, typically without end-to-end finetuning. For a block with weights $W$ and activations $X$, PTQ constructs quantized counterparts $(\hat W,\hat X)$. A common formulation fixes $\hat X=Q_X(X)$ during each block update and minimizes
\begin{equation}
\min_{\hat W}\;\mathbb E_{X\sim\mathcal D_{\rm calib}}\big[\|WX-\hat W\hat X\|_F^2\big],
\label{eq:ptq_recon}
\end{equation}
over quantization parameters for $\hat W$.

Eq.~\eqref{eq:ptq_recon} minimizes total feature error across all directions. SAM~3 also uses the relation between visual and text features to select a target. Perturbations with similar $L_2$ error can therefore produce different grounding results when they affect different feature directions. This motivates giving more calibration weight to directions that strongly affect concept scores.

\subsection{Prompt-Alignment Drift}
\label{subsec:empirical_observation}

SAM~3 combines mask prediction with prompt grounding~\cite{ravi2024sam,carion2025sam,ranjan2025mix,liu2024pq,wang2024q}. Under W4A8, reconstruction-based methods such as PTQ4SAM preserve much of the COCO/LVIS AP but lose accuracy on referring expressions and long video sequences. We use \emph{prompt-alignment drift} to describe these errors in target selection and identity propagation. Sec.~\ref{subsec:mechanism_checks} examines them through cross-task comparisons and matched feature interventions.

\subsection{Subspace Perturbation View}
\label{subsec:theoretical_analysis}

To motivate a geometric diagnostic, let $F\in\mathbb R^{N\times D}$ contain FP16 calibration tokens, let $F_c=F-\mathbf 1\mu^\top$, and define
\begin{equation}
\Sigma=\frac{1}{N}F_c^\top F_c\in\mathbb R^{D\times D}.
\end{equation}
Let $\Sigma=U\Lambda U^\top$ have eigenvalues $\lambda_1\ge\cdots\ge\lambda_D\ge0$. We partition
\begin{equation}
U=\begin{bmatrix}U_{\rm dom}&U_{\rm res}\end{bmatrix},
\end{equation}
where $U_{\rm dom}\in\mathbb R^{D\times k}$ contains the top-$k$ principal directions and $U_{\rm res}\in\mathbb R^{D\times r}$, $r=D-k$, is their low-energy complement. PCA selects these directions by variance; we measure their prompt sensitivity empirically. Quantization gives $\hat\Sigma=\Sigma+E$; Davis--Kahan bounds the rotation of the corresponding residual basis~\cite{davis1970rotation,yu2015useful}:
\begin{equation}
\|\sin\Theta(U_{\rm res},\hat U_{\rm res})\|_F
\le\frac{\|E\|_F}{\delta},\qquad
\delta=\lambda_k-\lambda_{k+1}.
\label{eq:davis_kahan}
\end{equation}
Eq.~\eqref{eq:davis_kahan} relates basis rotation to the covariance perturbation and the boundary eigengap. We assess grounding sensitivity through the controlled interventions in Supplementary Sec.~B. We select $k$ using concept-score drift on held-out calibration data and freeze it before evaluation. At the default $k/D=0.50$, the relative boundary gap has median $0.07$ (IQR $0.04$--$0.11$) across protected layers and calibration resamples. Rank sensitivity, a text-gradient alternative, and matched interventions are reported in Supplementary Sec.~B.

\section{Methodology}
\label{sec:method}

\begin{figure*}[t]
    \centering
    \includegraphics[width=\textwidth]{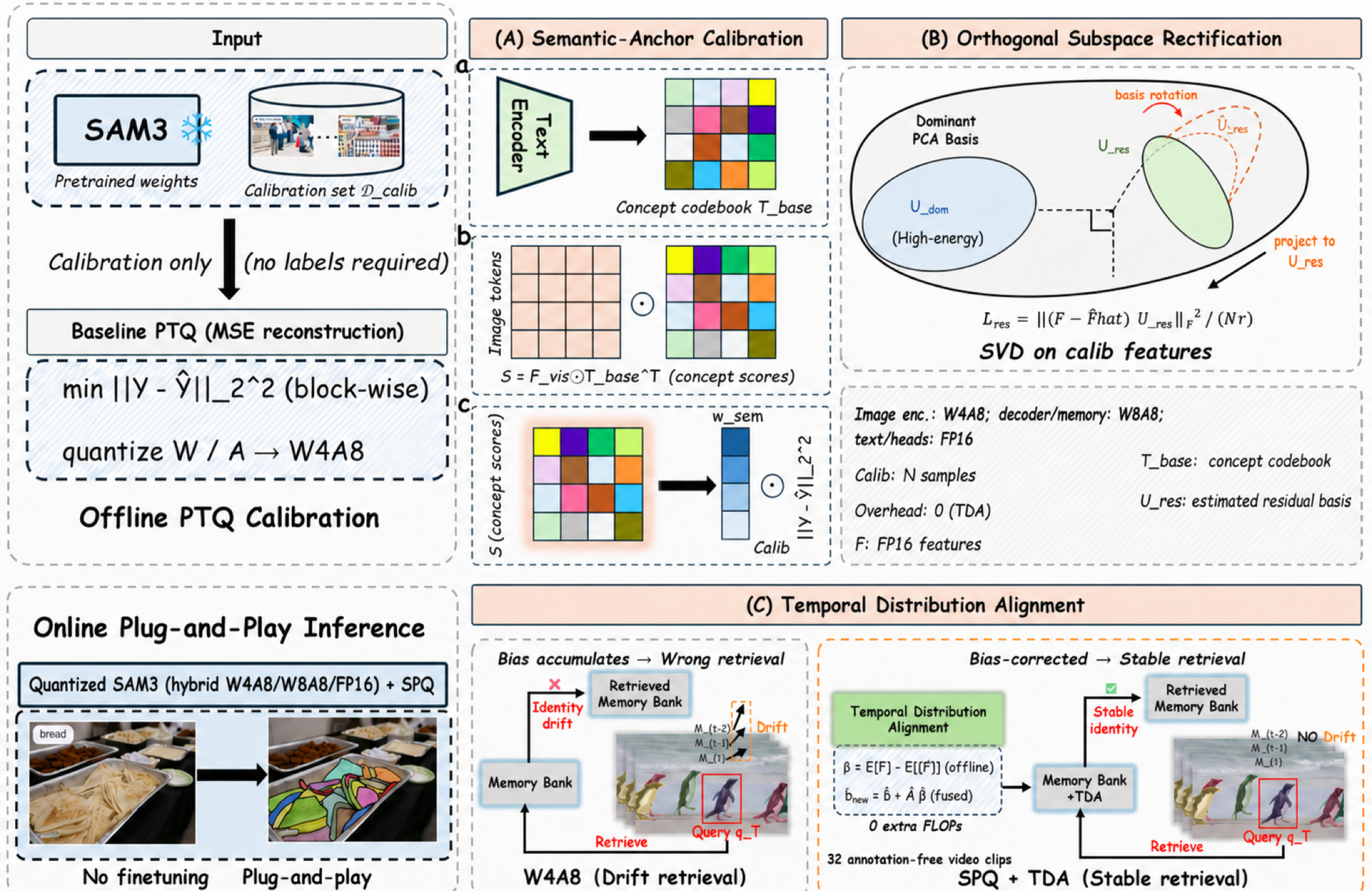}
    \caption{\textbf{Overview of SPQ.}
    Designs A and B calibrate image blocks with a fixed concept codebook and residual-basis feature matching. Design C estimates the mean memory error from paired video rollouts and folds the correction into an existing affine map. SPQ performs these steps offline and preserves the exported graph.}
    \label{fig:method_overview}
\end{figure*}

Figure~\ref{fig:method_overview} summarizes \textbf{Semantics-Preserving Quantization (SPQ)}. Design A uses concept scores to guide weight rounding. Design B preserves prompt-sensitive residual features. Design C corrects the mean error at recurrent memory interfaces.

\subsection{Semantic-Anchor Calibration}
\label{subsec:design_a}

\paragraph{Motivation.}
Standard PTQ calibration minimizes activation reconstruction error with equal weight across feature directions. SAM~3 also needs stable vision--language alignment to select the prompted object. Design A therefore gives greater weight to quantization decisions that affect cross-modal concept scores.

\paragraph{Concept codebook as semantic anchor.}
We build a fixed concept codebook to supply the text that SA-1B lacks. Qwen2.5-VL-72B-Instruct~\cite{bai2025qwen25vl} first generates short object-centric phrases from calibration images, using image content without human or evaluation annotations. We deduplicate the object/category names and add synonyms and fixed templates (e.g., ``a photo of \{concept\}''). The frozen SAM~3 text encoder embeds and normalizes each string. We then average template variants into one prototype in $T_{\text{base}}\in\mathbb R^{C\times D}$. We keep this codebook fixed across runs, reserve RefCOCO/+/g annotations for evaluation, and use the VLM offline. Supplementary Sec.~C reports vocabulary overlap and leave-concept-out transfer.

Given normalized visual features $F_{\text{vis}}\in\mathbb R^{B\times D}$, the concept-score matrix is
\begin{equation}
S=F_{\text{vis}}T_{\text{base}}^\top\in\mathbb R^{B\times C}.
\end{equation}

\paragraph{Concept-score curvature.}
Let $p=\dim(\operatorname{vec}(W))$. We use the Gauss--Newton-like concept-score curvature
\begin{equation}
\begin{aligned}
H_{\text{sem}}&=\mathbb E[J_W^\top J_W]\in\mathbb R^{p\times p},\\
J_W&=\frac{\partial\operatorname{vec}(S)}{\partial\operatorname{vec}(W)}
\in\mathbb R^{BC\times p}.
\end{aligned}
\label{eq:concept_fisher}
\end{equation}
This matrix measures the local sensitivity of concept scores to the block parameters. We compute a block-diagonal approximation through Jacobian--vector products, avoiding the cost of materializing the full matrix.

\paragraph{Semantics-aware reconstruction objective.}
With $\hat X=Q_X(X)$ fixed during a block update and $\Delta w=\operatorname{vec}(\hat W-W)$, Design A minimizes
\begin{equation}
\mathcal L_A=
\mathbb E\|WX-\hat W\hat X\|_F^2
+\alpha\,\Delta w^\top H_{\text{sem}}\Delta w,
\label{eq:semantic_anchor_obj}
\end{equation}
where $\alpha$ balances reconstruction and concept-score sensitivity. Design B complements this parameter-space objective by directly constraining the resulting feature error.

\subsection{Orthogonal Subspace Rectification}
\label{subsec:design_b}

\paragraph{Motivation.}
Low-bit quantization can shift residual feature directions even after concept-sensitive rounding. Design B penalizes these shifts at the feature level.

\paragraph{Estimating the residual basis.}
For centered calibration features $F_c\in\mathbb R^{N\times D}$, let $U_{\rm res}\in\mathbb R^{D\times r}$, $r=D-k$, be the complement of the top-$k$ PCA basis (Sec.~\ref{subsec:theoretical_analysis}). We choose $k$ using concept-score drift on held-out calibration data and freeze it before evaluation. Supplementary Sec.~B evaluates the selected basis with matched functional interventions.

\paragraph{Subspace-constrained loss.}
For matched FP16 and quantized layer features $F,\hat F\in\mathbb R^{N\times D}$, Design B and the final block objective are
\begin{equation}
\mathcal L_B=\frac{1}{Nr}\|(F-\hat F)U_{\rm res}\|_F^2,
\qquad
\mathcal L_{\rm SPQ}=\mathcal L_A+\lambda\mathcal L_B.
\label{eq:rectification_loss}
\end{equation}
The $Nr$ normalization keeps $\lambda$ comparable across layers and ranks. The loss directly reduces feature error along the selected residual coordinates.

\subsection{Temporal Distribution Alignment}
\label{subsec:design_c}

\paragraph{Motivation.}
SAM~3 uses a memory bank and memory cross-attention to propagate objects through video~\cite{ravi2024sam}. Small feature shifts can accumulate across frames, alter memory retrieval, and cause identity drift. Design C estimates and corrects the mean shift at the memory interface.

\paragraph{Offline bias shift estimation.}
Let $f_t^{\rm FP},f_t^{\rm Q}\in\mathbb R^{d_{\rm in}}$ be paired features at the same memory write/read interface, generated with identical clips, prompts, history initialization, and time indices. Per targeted layer or stage, we estimate
\begin{equation}
\beta=\mathbb E[f_t^{\rm FP}-f_t^{\rm Q}].
\label{eq:bias_shift}
\end{equation}
Every factorial cell and video baseline uses the same 32 annotation-free training clips, disjoint from SA-V validation/test. These paired video rollouts supply the memory statistics for Design C.

\paragraph{Zero-cost rectification via bias fusion.}
For a directly subsequent exported affine map $\hat y=\hat A f+\hat b$, when no normalization or nonlinearity intervenes, we export
\begin{equation}
\hat b_{\rm new}=\hat b+\hat A\beta.
\end{equation}
We estimate and project the correction separately for key and value paths. We fuse it into an existing bias or a backend-supported fused bias that preserves the graph. This closed-form correction targets the memory-feature means and preserves the deployed operators and FLOPs.

\section{Experiments}

\subsection{Experimental Setup}
\label{subsec:exp_setup}

We evaluate mask quality on COCO 2017~\cite{lin2014microsoft} and LVIS~\cite{gupta2019lvis}, prompt grounding on RefCOCO/+/g~\cite{yu2016modeling}, and video tracking on SA-V~\cite{ravi2024sam}. These tasks measure spatial segmentation, text-conditioned target selection, and temporal stability.

Unless stated otherwise, W4A8 uses 4-bit image-encoder weights and 8-bit activations. The multimodal decoder, detector queries, and memory use W8A8, while the text encoder and prediction heads remain FP16. In W8A8 rows, all quantized modules use 8-bit weights and activations, with the same FP16 components.

Designs A/B use 1,024 unlabeled SA-1B images~\cite{kirillov2023segment}. Design C and every video baseline use the same 32 annotation-free training clips, disjoint from SA-V validation/test. Paired rollouts share clips, prompts, history initialization, and time indices. We set $\alpha=0.1$, $\lambda=1$, and run 2,000 updates per block for A/B. Design C uses the closed-form estimate.

We adapt RepQ-ViT~\cite{li2023repq}, PTQ4SAM~\cite{lv2024ptq4sam}, SAQ-SAM~\cite{zhang2026saqsam}, and CAR-SAM~\cite{wen2026carsam} to the same checkpoint, calibration data, target modules, precision, and evaluation protocol. Mix-QSAM3~\cite{ranjan2026mixqsam3} retains its native mixed-precision allocation. Supplementary Sec.~D gives the full protocol and cross-task comparison. We measure all deployment metrics on the exported graph.

\subsection{Main Results on SAM 3}
\label{subsec:main_results}


\subsubsection{Structural Integrity on Segmentation}
\label{subsubsec:structural_integrity}

We first evaluate mask quality on COCO and LVIS (Table~\ref{tab:structural}). Under W4A8, Min-Max loses substantial AP, while PTQ4ViT and PTQ4SAM recover much of the loss. SPQ achieves the strongest AP among the reported PTQ baselines. These results establish the structural accuracy of the quantized checkpoints. We next test whether their text-conditioned grounding is equally well preserved.

\begin{table*}[!th]
\caption{\textbf{Structural Integrity on COCO and LVIS.} We report mask Average Precision (AP). The FP16 entries follow the official SAM 3 reported metrics. PTQ4SAM and SPQ both preserve much of the structural segmentation accuracy under low-bit quantization.}
\label{tab:structural}
\vspace{-8pt}
\centering
\setlength{\tabcolsep}{4pt}
\renewcommand{\arraystretch}{1.15} 

\resizebox{\textwidth}{!}{%
    \begin{tabular}{ll ccc cccc}
    \toprule
    \multirow{2}{*}{\textbf{Bit-width}} & \multirow{2}{*}{\textbf{Method}} & \multicolumn{3}{c}{\textbf{COCO Val}} & \multicolumn{4}{c}{\textbf{LVIS v1.0 Val}} \\
    
    \cmidrule(lr){3-5} \cmidrule(lr){6-9}
    & & \texttt{AP}~$\uparrow$
    & \texttt{AP}$_{50}$~$\uparrow$
    & \texttt{AP}$_{75}$~$\uparrow$
    & \texttt{AP}~$\uparrow$
    & \texttt{AP}$_{r}$~$\uparrow$
    & \texttt{AP}$_{c}$~$\uparrow$
    & \texttt{AP}$_{f}$~$\uparrow$ \\

    \midrule
    \multirow{1}{*}{FP16} & 
    \textcolor{gray}{Full Precision (Baseline)}
    & \textcolor{gray}{55.7} & \textcolor{gray}{76.2} & \textcolor{gray}{61.4} & \textcolor{gray}{48.5} & \textcolor{gray}{41.2} & \textcolor{gray}{47.5} & \textcolor{gray}{52.6} \\

    \midrule
    \multirow{4}{*}{\textbf{W8A8}} & 
    MinMax
    & 53.8 & 74.5 & 59.3 & 45.5 & 36.2 & 45.0 & 50.5 \\
    
    & PTQ4ViT~\cite{yuan2022ptq4vit}
    & 54.6$_{\textcolor{forestgreen}{(+0.8)}}$ & 75.4$_{\textcolor{forestgreen}{(+0.9)}}$ & 60.2$_{\textcolor{forestgreen}{(+0.9)}}$ & 46.8$_{\textcolor{forestgreen}{(+1.3)}}$ & 38.1$_{\textcolor{forestgreen}{(+1.9)}}$ & 46.1$_{\textcolor{forestgreen}{(+1.1)}}$ & 51.5 \\

    & PTQ4SAM~\cite{lv2024ptq4sam}
    & 55.2$_{\textcolor{forestgreen}{(+1.4)}}$ & 75.8$_{\textcolor{forestgreen}{(+1.3)}}$ & 60.9$_{\textcolor{forestgreen}{(+1.6)}}$ & 47.9$_{\textcolor{forestgreen}{(+2.4)}}$ & 39.6$_{\textcolor{forestgreen}{(+3.4)}}$ & 46.9$_{\textcolor{forestgreen}{(+1.9)}}$ & 52.0 \\

    & \cellcolor{lightblue!60} \textbf{SPQ (Ours)}
    & \cellcolor{lightblue!60}\textbf{55.6}$_{\textcolor{forestgreen}{(+1.8)}}$ 
    & \cellcolor{lightblue!60}\textbf{76.1}$_{\textcolor{forestgreen}{(+1.6)}}$ 
    & \cellcolor{lightblue!60}\textbf{61.2}$_{\textcolor{forestgreen}{(+1.9)}}$ 
    & \cellcolor{lightblue!60}\textbf{48.4}$_{\textcolor{forestgreen}{(+2.9)}}$ 
    & \cellcolor{lightblue!60}\textbf{40.5}$_{\textcolor{forestgreen}{(+4.3)}}$ 
    & \cellcolor{lightblue!60}\textbf{47.3}$_{\textcolor{forestgreen}{(+2.3)}}$ 
    & \cellcolor{lightblue!60}\textbf{52.4} \\ 

    \midrule[\heavyrulewidth] 

    \multirow{4}{*}{\textbf{W4A8}} & 
    MinMax 
    & 45.2 & 66.4 & 47.6 & 38.5 & 28.2 & 37.4 & 43.1 \\

    & PTQ4ViT~\cite{yuan2022ptq4vit}
    & 50.5$_{\textcolor{forestgreen}{(+5.3)}}$ & 72.1$_{\textcolor{forestgreen}{(+5.7)}}$ & 54.2$_{\textcolor{forestgreen}{(+6.6)}}$ & 44.6$_{\textcolor{forestgreen}{(+6.1)}}$ & 36.2$_{\textcolor{forestgreen}{(+8.0)}}$ & 43.8$_{\textcolor{forestgreen}{(+6.4)}}$ & 48.8 \\

    & PTQ4SAM~\cite{lv2024ptq4sam}
    & 54.1$_{\textcolor{forestgreen}{(+8.9)}}$ & 75.0$_{\textcolor{forestgreen}{(+8.6)}}$ & 59.8$_{\textcolor{forestgreen}{(+12.2)}}$ & 46.8$_{\textcolor{forestgreen}{(+8.3)}}$ & 38.8$_{\textcolor{forestgreen}{(+10.6)}}$ & 45.5$_{\textcolor{forestgreen}{(+8.1)}}$ & 50.7 \\

    & \cellcolor{lightblue!60} \textbf{SPQ (Ours)}
    & \cellcolor{lightblue!60}\textbf{54.8}$_{\textcolor{forestgreen}{(+9.6)}}$ 
    & \cellcolor{lightblue!60}\textbf{75.6}$_{\textcolor{forestgreen}{(+9.2)}}$ 
    & \cellcolor{lightblue!60}\textbf{60.6}$_{\textcolor{forestgreen}{(+13.0)}}$ 
    & \cellcolor{lightblue!60}\textbf{47.5}$_{\textcolor{forestgreen}{(+9.0)}}$ 
    & \cellcolor{lightblue!60}\textbf{39.7}$_{\textcolor{forestgreen}{(+11.5)}}$ 
    & \cellcolor{lightblue!60}\textbf{46.8}$_{\textcolor{forestgreen}{(+9.4)}}$ 
    & \cellcolor{lightblue!60}\textbf{51.8} \\ 

    \bottomrule
    \end{tabular}
}
\vspace{-8pt}
\end{table*}

\subsubsection{Zero-Shot Concept Alignment}
\label{subsubsec:concept_alignment}

Table~\ref{tab:concept} reports text-guided grounding on RefCOCO/+/g. PTQ4SAM loses much more accuracy on referring expressions than on structural AP. SPQ recovers much of this grounding gap under W4A8 and remains close to FP16 under W8A8. Matched interventions in Supplementary Sec.~B further test how grounding responds to changes in the residual basis.

\begin{table*}[!th]
\caption{\textbf{Zero-shot referring segmentation.} Reconstruction PTQ loses substantially more prompt-grounding accuracy than structural AP.}
\label{tab:concept}
\vspace{-8pt}
\centering
\setlength{\tabcolsep}{4pt}
\renewcommand{\arraystretch}{1.15} 

\resizebox{\textwidth}{!}{%
    \begin{tabular}{ll ccc ccc}
    \toprule
    \multirow{2}{*}{\textbf{Bit-width}} & \multirow{2}{*}{\textbf{Method}} & \multicolumn{2}{c}{\textbf{RefCOCO}} & \multicolumn{2}{c}{\textbf{RefCOCO+}} & \multicolumn{2}{c}{\textbf{RefCOCOg}} \\
    
    \cmidrule(lr){3-4} \cmidrule(lr){5-6} \cmidrule(lr){7-8}
    & & \texttt{Pr@0.5}~$\uparrow$
    & \texttt{oIoU}~$\uparrow$
    & \texttt{Pr@0.5}~$\uparrow$
    & \texttt{oIoU}~$\uparrow$
    & \texttt{Pr@0.5}~$\uparrow$
    & \texttt{oIoU}~$\uparrow$ \\

    \midrule
    \multirow{1}{*}{FP16} & 
    \textcolor{gray}{Full Precision (Baseline)}
    & \textcolor{gray}{75.5} & \textcolor{gray}{66.8} & \textcolor{gray}{67.3} & \textcolor{gray}{59.5} & \textcolor{gray}{73.4} & \textcolor{gray}{65.2} \\

    \midrule
    \multirow{4}{*}{\textbf{W8A8}} & 
    MinMax
    & 61.2 & 53.1 & 52.5 & 45.2 & 59.1 & 51.5 \\

    & PTQ4ViT~\cite{yuan2022ptq4vit}
    & 63.5$_{\textcolor{forestgreen}{(+2.3)}}$ & 55.4$_{\textcolor{forestgreen}{(+2.3)}}$ & 55.2$_{\textcolor{forestgreen}{(+2.7)}}$ & 48.6$_{\textcolor{forestgreen}{(+3.4)}}$ & 61.2$_{\textcolor{forestgreen}{(+2.1)}}$ & 54.8$_{\textcolor{forestgreen}{(+3.3)}}$ \\

    & PTQ4SAM~\cite{lv2024ptq4sam}
    & 65.5$_{\textcolor{forestgreen}{(+4.3)}}$ & 57.2$_{\textcolor{forestgreen}{(+4.1)}}$ & 57.8$_{\textcolor{forestgreen}{(+5.3)}}$ & 50.1$_{\textcolor{forestgreen}{(+4.9)}}$ & 63.5$_{\textcolor{forestgreen}{(+4.4)}}$ & 56.5$_{\textcolor{forestgreen}{(+5.0)}}$ \\

    & \cellcolor{lightblue!60} \textbf{SPQ (Ours)}
    & \cellcolor{lightblue!60}\textbf{74.8}$_{\textcolor{forestgreen}{(+13.6)}}$ 
    & \cellcolor{lightblue!60}\textbf{66.5}$_{\textcolor{forestgreen}{(+13.4)}}$ 
    & \cellcolor{lightblue!60}\textbf{66.8}$_{\textcolor{forestgreen}{(+14.3)}}$ 
    & \cellcolor{lightblue!60}\textbf{58.8}$_{\textcolor{forestgreen}{(+13.6)}}$ 
    & \cellcolor{lightblue!60}\textbf{72.8}$_{\textcolor{forestgreen}{(+13.7)}}$ 
    & \cellcolor{lightblue!60}\textbf{64.9}$_{\textcolor{forestgreen}{(+13.4)}}$ \\ 

    \midrule[\heavyrulewidth] 

    \multirow{4}{*}{\textbf{W4A8}} & 
    MinMax 
    & 31.1 & 24.5 & 25.4 & 18.1 & 29.5 & 22.8 \\

    & PTQ4ViT~\cite{yuan2022ptq4vit}
    & 38.8$_{\textcolor{forestgreen}{(+7.7)}}$ & 31.6$_{\textcolor{forestgreen}{(+7.1)}}$ & 32.1$_{\textcolor{forestgreen}{(+6.7)}}$ & 24.8$_{\textcolor{forestgreen}{(+6.7)}}$ & 36.5$_{\textcolor{forestgreen}{(+7.0)}}$ & 28.5$_{\textcolor{forestgreen}{(+5.7)}}$ \\

    & PTQ4SAM~\cite{lv2024ptq4sam}
    & 42.1$_{\textcolor{forestgreen}{(+11.0)}}$ & 35.4$_{\textcolor{forestgreen}{(+10.9)}}$ & 36.5$_{\textcolor{forestgreen}{(+11.1)}}$ & 29.2$_{\textcolor{forestgreen}{(+11.1)}}$ & 40.2$_{\textcolor{forestgreen}{(+10.7)}}$ & 32.8$_{\textcolor{forestgreen}{(+10.0)}}$ \\

    & \cellcolor{lightblue!60} \textbf{SPQ (Ours)}
    & \cellcolor{lightblue!60}\textbf{72.5}$_{\textcolor{forestgreen}{(+41.4)}}$ 
    & \cellcolor{lightblue!60}\textbf{64.2}$_{\textcolor{forestgreen}{(+39.7)}}$ 
    & \cellcolor{lightblue!60}\textbf{65.1}$_{\textcolor{forestgreen}{(+39.7)}}$ 
    & \cellcolor{lightblue!60}\textbf{57.2}$_{\textcolor{forestgreen}{(+39.1)}}$ 
    & \cellcolor{lightblue!60}\textbf{71.0}$_{\textcolor{forestgreen}{(+41.5)}}$ 
    & \cellcolor{lightblue!60}\textbf{62.5}$_{\textcolor{forestgreen}{(+39.7)}}$ \\ 

    \bottomrule
    \end{tabular}
}
\vspace{-8pt}
\end{table*}

\begin{figure*}[t]
    \centering
    \includegraphics[width=0.96\textwidth]{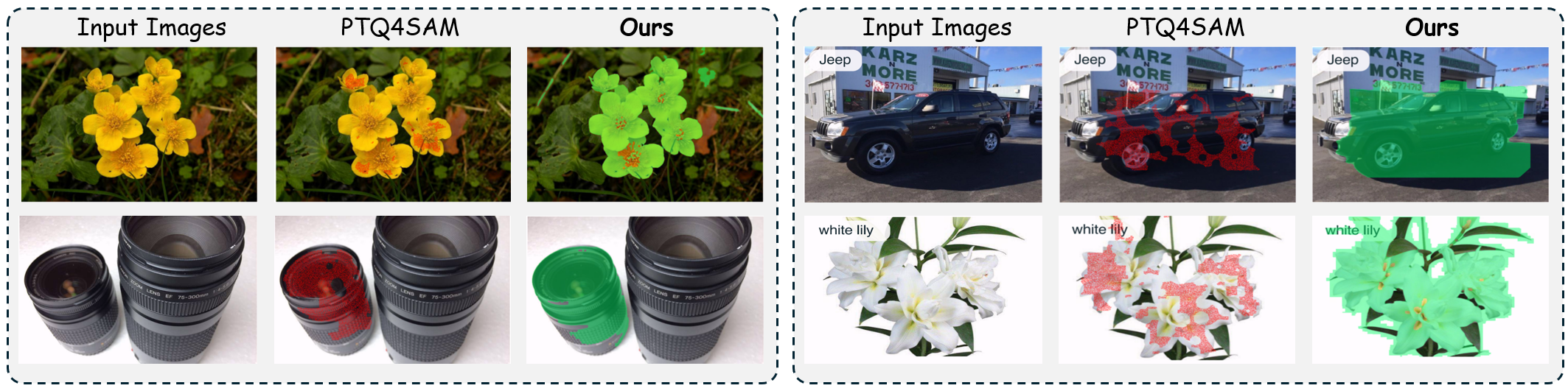}\\[-1mm]
    \includegraphics[width=0.96\textwidth]{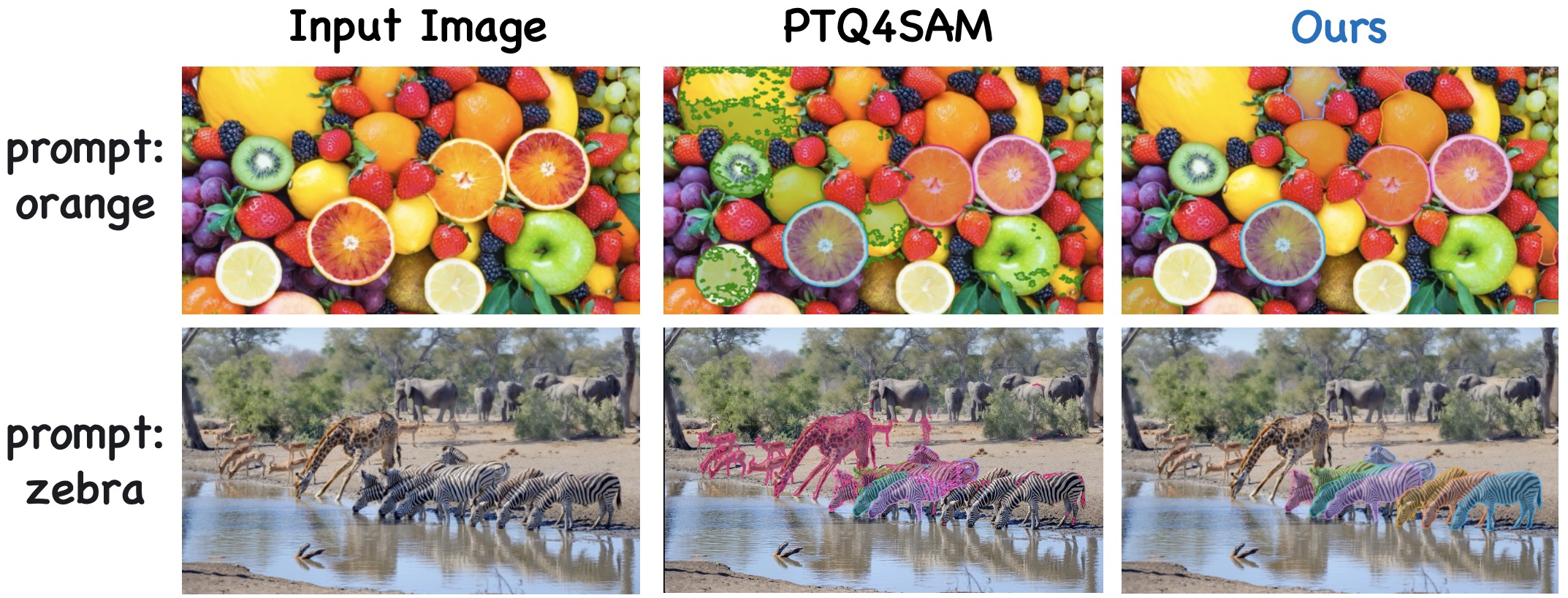}
    \caption{\textbf{Qualitative comparison under W4A8.}
    \textbf{Top:} instance-segmentation and text-prompted segmentation cases, with each text prompt shown in its input image.
    \textbf{Bottom:} additional prompt-labeled cases, including ``orange'' and ``zebra,'' that isolate prompt-conditioned target selection.
    SPQ selects the intended objects and returns more coherent masks than PTQ4SAM in the shown examples.}
    \label{fig:prompt_qualitative}
    \vspace{-8pt}
\end{figure*}

\textbf{Qualitative Comparison.}
Fig.~\ref{fig:prompt_qualitative} shows two W4A8 failure modes: selecting the wrong object and returning an incomplete mask. SPQ produces more coherent masks for the intended objects in these examples, consistent with its higher RefCOCO scores.

\subsubsection{Temporal Stability in Video Tracking}
\label{subsubsec:temporal_stability}

\begin{figure*}[t]
    \centering
    \includegraphics[width=0.96\textwidth]{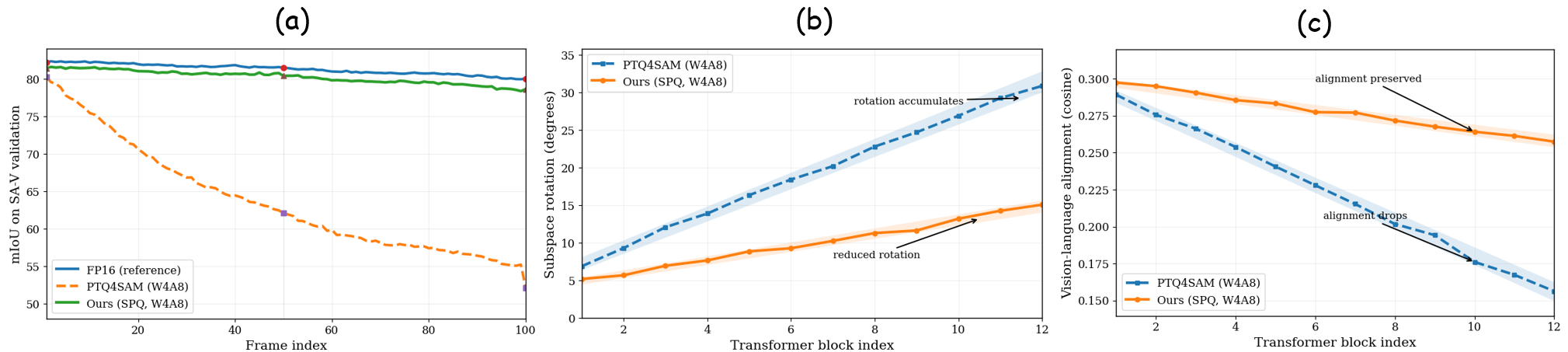}
    \caption{\textbf{Temporal degradation and prompt-alignment diagnostics.}
    (a) SA-V long-horizon tracking mIoU: PTQ4SAM under W4A8 decays over time, while SPQ remains close to FP16.
    (b) Angle between the estimated FP16 and quantized residual bases across transformer blocks: standard PTQ shows a larger angle, while SPQ shows a smaller one.
    (c) Vision-language alignment across blocks: PTQ4SAM drops with depth, while SPQ better preserves alignment. Shaded regions denote sample variability.}
    \label{fig:temp_rot_align}
    \vspace{-8pt}
\end{figure*}

We next evaluate SA-V tracking to measure how quantization error changes over time. Table~\ref{tab:temporal} reports early-frame quality and long-horizon decay. TDA uses the paired 32-clip protocol above and folds its mean correction into existing projections. Fig.~\ref{fig:temp_rot_align} describes the accompanying changes in tracking accuracy, residual-basis rotation, and vision--language alignment.

\begin{table*}[!th]
\caption{\textbf{Temporal stability on SA-V.} Progressive mIoU, J\&F, and consistency under the shared 32-clip calibration protocol.}
\label{tab:temporal}
\vspace{-8pt}
\centering
\setlength{\tabcolsep}{5pt}
\renewcommand{\arraystretch}{1.15}

\resizebox{\textwidth}{!}{%
    \begin{tabular}{ll ccc cc}
    \toprule
    \multirow{2}{*}{\textbf{Bit-width}} & \multirow{2}{*}{\textbf{Method}} & \multicolumn{3}{c}{\textbf{Progressive Frame mIoU}} & \multirow{2}{*}{\textbf{Video J\&F}~$\uparrow$} & \multirow{2}{*}{\textbf{Consistency Score ($\mathcal{T}_{cs}$)}~$\uparrow$} \\
    
    \cmidrule(lr){3-5} 
    & & \texttt{Frame 1}~$\uparrow$
    & \texttt{Frame 50}~$\uparrow$
    & \texttt{Frame 100}~$\uparrow$
    & & \\

    \midrule
    \multirow{1}{*}{FP16} & 
    \textcolor{gray}{Full Precision (Baseline)}
    & \textcolor{gray}{85.0} & \textcolor{gray}{83.5} & \textcolor{gray}{82.2} & \textcolor{gray}{83.5} & \textcolor{gray}{96.7\%} \\

    \midrule
    \multirow{4}{*}{\textbf{W8A8}} & 
    MinMax
    & 80.1 & 72.5 & 63.8 & 71.5 & 79.6\% \\
    
    & PTQ4ViT~\cite{yuan2022ptq4vit}
    & 82.2$_{\textcolor{forestgreen}{(+2.1)}}$ & 76.5$_{\textcolor{forestgreen}{(+4.0)}}$ & 69.8$_{\textcolor{forestgreen}{(+6.0)}}$ & 76.2$_{\textcolor{forestgreen}{(+4.7)}}$ & 84.9\% \\

    & PTQ4SAM~\cite{lv2024ptq4sam}
    & 83.6$_{\textcolor{forestgreen}{(+3.5)}}$ & 78.8$_{\textcolor{forestgreen}{(+6.3)}}$ & 72.5$_{\textcolor{forestgreen}{(+8.7)}}$ & 78.5$_{\textcolor{forestgreen}{(+7.0)}}$ & 86.7\% \\

    & \cellcolor{lightblue!60} \textbf{SPQ (Ours)}
    & \cellcolor{lightblue!60}\textbf{84.2}$_{\textcolor{forestgreen}{(+4.1)}}$ 
    & \cellcolor{lightblue!60}\textbf{83.1}$_{\textcolor{forestgreen}{(+10.6)}}$ 
    & \cellcolor{lightblue!60}\textbf{82.2}$_{\textcolor{forestgreen}{(+18.4)}}$ 
    & \cellcolor{lightblue!60}\textbf{83.2}$_{\textcolor{forestgreen}{(+11.7)}}$ 
    & \cellcolor{lightblue!60}\textbf{97.6\%} \\ 

    \midrule[\heavyrulewidth] 

    \multirow{4}{*}{\textbf{W4A8}} & 
    MinMax 
    & 74.4 & 54.2 & 41.1 & 56.5 & 55.2\% \\

    & PTQ4ViT~\cite{yuan2022ptq4vit}
    & 78.5$_{\textcolor{forestgreen}{(+4.1)}}$ & 62.8$_{\textcolor{forestgreen}{(+8.6)}}$ & 48.5$_{\textcolor{forestgreen}{(+7.4)}}$ & 63.4$_{\textcolor{forestgreen}{(+6.9)}}$ & 61.7\% \\

    & PTQ4SAM~\cite{lv2024ptq4sam}
    & 80.2$_{\textcolor{forestgreen}{(+5.8)}}$ & 66.5$_{\textcolor{forestgreen}{(+12.3)}}$ & 52.1$_{\textcolor{forestgreen}{(+11.0)}}$ & 65.4$_{\textcolor{forestgreen}{(+8.9)}}$ & 65.0\% \\

    & \cellcolor{lightblue!60} \textbf{SPQ (Ours)}
    & \cellcolor{lightblue!60}\textbf{83.5}$_{\textcolor{forestgreen}{(+9.1)}}$ 
    & \cellcolor{lightblue!60}\textbf{81.5}$_{\textcolor{forestgreen}{(+27.3)}}$ 
    & \cellcolor{lightblue!60}\textbf{80.1}$_{\textcolor{forestgreen}{(+39.0)}}$ 
    & \cellcolor{lightblue!60}\textbf{81.2}$_{\textcolor{forestgreen}{(+24.7)}}$ 
    & \cellcolor{lightblue!60}\textbf{95.9\%} \\ 

    \bottomrule
    \end{tabular}
}
\vspace{-8pt}
\end{table*}

\begin{table*}[!th]
\caption{\textbf{Complete $2^3$ factorial under W4A8.} All cells use identical precision and calibration data.}
\label{tab:ablation}
\vspace{-8pt}
\centering
\setlength{\tabcolsep}{8pt}
\renewcommand{\arraystretch}{1.06}

\resizebox{0.95\textwidth}{!}{%
    \begin{tabular}{ccc | ccc}
    \toprule
    \textbf{Design A} & \textbf{Design B} & \textbf{Design C} & \textbf{Structure} & \textbf{Grounding} & \textbf{Temporal} \\
    (Concept Calib.) & (Residual Rect.) & (Temporal Align.) & COCO AP $\uparrow$ & RefCOCO Pr@0.5 $\uparrow$ & SA-V J\&F $\uparrow$ \\
    \midrule
    
    \textcolor{gray}{\ding{55}} & \textcolor{gray}{\ding{55}} & \textcolor{gray}{\ding{55}} & 
    45.2 & 31.1 & 56.5 \\
    
    \checkmark & \textcolor{gray}{\ding{55}} & \textcolor{gray}{\ding{55}} & 52.5 & 42.8 & 62.1 \\
    \textcolor{gray}{\ding{55}} & \checkmark & \textcolor{gray}{\ding{55}} & 47.1 & 65.9 & 58.4 \\
    \textcolor{gray}{\ding{55}} & \textcolor{gray}{\ding{55}} & \checkmark & 45.4 & 31.5 & 75.6 \\
    \checkmark & \checkmark & \textcolor{gray}{\ding{55}} & 54.5 & 72.1 & 68.4 \\
    \checkmark & \textcolor{gray}{\ding{55}} & \checkmark & 52.7 & 43.1 & 79.3 \\
    \textcolor{gray}{\ding{55}} & \checkmark & \checkmark & 47.4 & 66.3 & 78.8 \\
    \rowcolor{lightblue!60}
    \checkmark & \checkmark & \checkmark & 
    \textbf{54.8} & \textbf{72.5} & \textbf{81.2} \\
    
    \bottomrule
    \end{tabular}
}
\vspace{-8pt}
\end{table*}

\subsection{Ablation Study}
\label{subsec:ablation}

Table~\ref{tab:ablation} evaluates all component combinations under the same W4A8 protocol. A alone raises COCO AP by 7.3, B alone raises RefCOCO Pr@0.5 by 34.8, and C alone raises SA-V J\&F by 19.1 over MinMax. The full model combines these gains and reaches 54.8/72.5/81.2. C acts on the video graph, so we treat the paired rows' 0.2--0.4 differences on COCO/RefCOCO as run variation. Supplementary Sec.~B examines the cross-task effects of the feature interventions.

\paragraph{Additional experiments.}
Sec.~\ref{subsec:mechanism_checks} summarizes the method comparisons, functional controls, codebook analysis, calibration-scale study, and temporal results. Supplementary Secs.~B--E give complete tables and protocols; Secs.~F--I cover deployment accounting, efficiency, and transfer. Figs.~\ref{fig:prompt_qualitative} and~\ref{fig:temp_rot_align} show the qualitative and temporal results, respectively.

\subsection{Cross-Model Transfer}
\label{subsec:clip_transfer}

\begin{table}[!ht]
\caption{\textbf{CLIP ViT-B/16 transfer under W4A8.} Zero-shot R@1; the full R@1/R@5 table is in Supplementary Sec.~I.}
\label{tab:clip_transfer_main}
\vspace{-5pt}
\centering
\scriptsize
\setlength{\tabcolsep}{3.1pt}
\begin{tabular}{lcccc}
\toprule
& \multicolumn{2}{c}{Flickr30k} & \multicolumn{2}{c}{MSCOCO} \\
\cmidrule(lr){2-3}\cmidrule(lr){4-5}
Method & I2T & T2I & I2T & T2I \\
\midrule
FP16 & 81.2 & 62.4 & 58.2 & 38.5 \\
MinMax & 42.5 & 31.2 & 28.5 & 18.2 \\
PTQ4ViT~\cite{yuan2022ptq4vit} & 58.4 & 44.8 & 41.2 & 26.4 \\
\rowcolor{lightblue!60}\textbf{SPQ} & \textbf{78.5} & \textbf{59.8} & \textbf{55.8} & \textbf{36.7} \\
\bottomrule
\end{tabular}
\vspace{-6pt}
\end{table}

Applying Designs A+B to CLIP recovers substantially more zero-shot image--text retrieval accuracy than reconstruction PTQ on both datasets. The gains extend the calibration strategy to a separate vision--language model.

\subsection{Mechanism, Calibration, and Scope}
\label{subsec:mechanism_checks}

\paragraph{Task-specific degradation and functional controls.}
Under W4A8, PTQ4SAM is $1.6$ points below FP16 on COCO AP ($54.1$ versus $55.7$) and $33.4$ points below on RefCOCO Pr@0.5 ($42.1$ versus $75.5$). SPQ reaches $54.8/72.5$. To examine this task asymmetry, we apply matched-rank, matched-norm interventions to the same unperturbed quantized checkpoint. Perturbing $U_{\rm res}$ gives $\Delta$AP/$\Delta$Ref of $-1.3/-22.4$, compared with $-8.7/-4.6$ for $U_{\rm dom}$ and $-3.0/-7.2$ for a rank-matched random basis. Removing the two bases gives $-2.1/-26.8$ and $-10.2/-5.1$, respectively. Residual-basis changes affect grounding most strongly, while dominant-basis changes affect AP most strongly. Both bases affect both tasks, showing selective sensitivity with cross-effects (Supplementary Sec.~B).

\paragraph{Rank selection and basis choice.}
We select rank using concept-score drift on held-out calibration data and freeze it before evaluation, with RefCOCO reserved for evaluation. For $k/D=\{0.25,0.50,0.75\}$, COCO AP/RefCOCO Pr@0.5 is $54.4/71.8$, $54.8/72.5$, and $54.6/71.9$. An equal-rank basis of held-out visual--text score gradients gives $54.7/72.8$. Performance remains similar across these tested ranks and basis constructions. Sec.~\ref{subsec:theoretical_analysis} provides the complementary eigengap diagnostic.

\paragraph{Calibration coverage and sample efficiency.}
We build the codebook offline and keep it fixed across runs. RefCOCO/+/g annotations are reserved for evaluation, and the VLM is used for offline codebook construction. Exact/lemmatized head-noun overlap is $48.2/61.3\%$, $45.6/58.7\%$, and $46.9/60.1\%$ on RefCOCO, RefCOCO+, and RefCOCOg. Codebook-present/absent concepts obtain $73.0/69.4$ RefCOCO Pr@0.5; leave-concept-out/full-codebook calibration gives $70.6/72.5$. SPQ therefore transfers beyond exact codebook entries, with a measurable gain from broader vocabulary coverage. Using 32 A/B images gives RefCOCO Pr@0.5/oIoU and SA-V J\&F of $70.2/62.1/78.8$, compared with $72.5/64.2/81.2$ using 1,024 images. Every setting uses the same 32 TDA clips, so this comparison measures image-calibration scale under fixed video data (Supplementary Secs.~C and E).

\paragraph{Controlled closest-method comparison.}
The adapted methods share the checkpoint, calibration data, target modules, hybrid W4A8 assignment, and evaluation protocol. Each uses its published calibration rule. COCO/RefCOCO/SA-V scores are $51.7/40.8/64.0$ for RepQ-ViT, $54.3/58.9/66.8$ for SAQ-SAM, $54.5/61.7/69.4$ for CAR-SAM, and $54.8/72.5/81.2$ for SPQ. Mix-QSAM3 reaches $54.4/67.8/79.0$ under its native mixed-precision allocation, providing an additional accuracy comparison at that deployment setting. We discuss Q-VLM and VLMQ as related work because adapting them requires a SAM~3 quantization protocol. Together, these results evaluate SPQ's combination of calibration objectives and memory correction on SAM~3 (Supplementary Sec.~D).

\paragraph{Temporal and factorial evidence.}
Under the shared 32-clip protocol, PTQ4SAM loses $28.1$ mIoU points from Frame~1 to Frame~100 ($80.2$ to $52.1$). SPQ loses $3.4$ ($83.5$ to $80.1$), with consistency improving from $65.0\%$ for PTQ4SAM to $95.9\%$ for SPQ. The accuracy gap grows over time, highlighting the importance of stable memory propagation. Table~\ref{tab:ablation} shows how the components work together. Adding B to A raises grounding from $42.8$ to $72.1$; adding A to B raises COCO AP from $47.1$ to $54.5$ and grounding from $65.9$ to $72.1$. Adding C to A+B raises SA-V J\&F from $68.4$ to $81.2$, alongside $0.3/0.4$ static-metric variation. Secs.~\ref{subsec:exp_setup} and~\ref{subsec:design_c} specify the paired-rollout protocol and bias-fusion conditions.

\paragraph{Efficiency and deployment accounting.}
On an RTX~3090 with $1024\!\times\!1024$ video frames, exported W4A8 reduces storage from $1.35$ to $0.38$~GB, peak memory from $14.8$ to $7.9$~GB, and latency from $42.5$ to $18.2$~ms/frame. Throughput rises from $23.5$ to $54.9$~FPS. At W8A8, SPQ and PTQ4SAM both use $0.68$~GB storage, $8.5$~GB memory, $25.4$~ms/frame, and $39.3$~FPS. Reduced precision provides the runtime gains; offline A/B calibration and bias-fused C improve accuracy on the same graph. Inference latency measures the exported model, with concept scoring, curvature approximation, basis estimation, and video rollouts accounted for as one-time calibration work. These measurements characterize the hybrid deployment, which keeps decoder/memory at 8 bits and text components at FP16 (Supplementary Secs.~F and H).

\paragraph{Sensitivity, qualitative evidence, and scope.}
Supplementary Sec.~A reports the full sensitivity sweep. RefCOCO Pr@0.5 is $68.4/72.5/71.9$ for $\alpha=0.01/0.1/1.0$ at $\lambda=1$, and $65.2/72.5/72.3$ for $\lambda=0.1/1/10$ at $\alpha=0.1$. Performance remains stable for $\lambda\in[1,10]$, while excessive concept weighting reduces structural AP. We fix $(\alpha,\lambda)=(0.1,1)$ before evaluation. The prompt-labeled examples illustrate wrong-target selection and incomplete masks, complementing the quantitative geometry and grounding measurements. The CLIP experiment extends A+B to one additional model. Deployment vocabulary shifts may require a new codebook. Broader architectures, lower bit-widths, and a more complete accuracy--cost evaluation remain future work.

\section{Conclusion}
\label{sec:conclusion}

SPQ combines concept-sensitive calibration, residual-feature matching, and a correction for video memory. Controlled interventions show that grounding is especially sensitive to errors in low-variance residual directions. Experiments show gains in image segmentation, referring segmentation, and video tracking with the inference graph unchanged. Future work will extend SPQ to more architectures and lower activation bit widths.

\section*{Acknowledgments}
This manuscript was co-authored by Oak Ridge National Laboratory (ORNL), operated by UT-Battelle, LLC under Contract No. DE-AC05-00OR22725 with the U.S. Department of Energy.
Any subjective views or opinions expressed in this paper do not necessarily represent those of the U.S. Department of Energy or the United States Government.

\clearpage
\twocolumn[{\centering\Large\bfseries Supplementary Material\par\vspace{1em}}]
\renewcommand{\thefigure}{S\arabic{figure}}
\renewcommand{\theHfigure}{S\arabic{figure}}
\setcounter{figure}{0}
\renewcommand{\thetable}{S\arabic{table}}
\renewcommand{\theHtable}{S\arabic{table}}
\setcounter{table}{0}

\appendix

\section*{Supplementary Roadmap}

\noindent This supplementary material is organized as follows.
\vspace{0.4em}

\begin{center}
\small
\setlength{\tabcolsep}{4pt}
\renewcommand{\arraystretch}{1.12}
\begin{tabular}{@{}lp{0.74\linewidth}@{}}
\toprule
\textbf{Sec.} & \textbf{Content} \\
\midrule
A & Hyper-parameter sensitivity for $\alpha$ and $\lambda$. \\
B & Residual-basis validation, rank selection, and functional controls. \\
C & Calibration-text construction and concept-transfer diagnostics. \\
D & Controlled comparison with the closest adapted methods. \\
E & Image-calibration size and fixed-video TDA protocol. \\
F & Lower-bit scope and offline calibration accounting. \\
G & Discussion of the temporal diagnostics shown in the main paper. \\
H & Hardware efficiency on the exported quantized graph. \\
I & CLIP transfer study for broader vision-language alignment. \\
J & Discussion of the qualitative comparisons shown in the main paper. \\
K & Related work. \\
L & Future work and limitations. \\
M & LLM usage statement. \\
\bottomrule
\end{tabular}
\end{center}

\vspace{0.3em}

\section{Hyper-parameter Sensitivity Analysis}
\label{supp:sensitivity}

We vary the concept-score curvature weight $\alpha$ in Design~A and the residual penalty strength $\lambda$ in Design~B. We measure COCO mask AP and RefCOCO grounding under W4A8 to examine how each weight affects structural and prompt-conditioned accuracy.

\begin{table}[h]
\caption{\textbf{Sensitivity Analysis of $\alpha$ and $\lambda$ on RefCOCO.} The combination $\alpha=0.1$ and $\lambda=1.0$ achieves the highest grounding accuracy among these settings.}
\label{tab:supp_sensitivity}
\centering
\small
\setlength{\tabcolsep}{10pt}
\resizebox{\linewidth}{!}{%
\begin{tabular}{ccccc}
\toprule
$\alpha$ & $\lambda$ & COCO AP $\uparrow$ & RefCOCO Pr@0.5 $\uparrow$ & RefCOCO oIoU $\uparrow$ \\
\midrule
0.01 & 1.0 & 54.7 & 68.4 & 60.1 \\
0.1  & 1.0 & \textbf{54.8} & \textbf{72.5} & \textbf{64.2} \\
1.0  & 1.0 & 53.2 & 71.9 & 63.8 \\
\midrule
0.1  & 0.1 & 54.8 & 65.2 & 57.4 \\
0.1  & 1.0 & \textbf{54.8} & \textbf{72.5} & \textbf{64.2} \\
0.1  & 10.0 & 54.5 & 72.3 & 64.0 \\
\bottomrule
\end{tabular}
}
\end{table}

\textbf{Analysis.} Performance changes little over $\lambda\in[1,10]$ in Table~\ref{tab:supp_sensitivity}. Increasing $\alpha$ to $1.0$ lowers COCO AP, showing the trade-off between emphasizing concept scores and preserving reconstruction accuracy.

\section{Residual-Basis Validation and Rank Selection}
\label{supp:basis_validation}

PCA orders feature directions by variance. We denote the top-$k$ dominant basis by $U_{\rm dom}$ and its low-energy complement by $U_{\rm res}$. We estimate prompt sensitivity through the functional interventions below. We select $k$ using concept-score drift on held-out calibration data and freeze it before RefCOCO evaluation. At the default $k/D=0.50$, the relative boundary eigengap
$(\lambda_k-\lambda_{k+1})/(\lambda_k+\epsilon)$, aggregated over protected layers and calibration resamples, has median $0.07$ (IQR $0.04$--$0.11$).
For the text-anchored alternative, we compute $g_c=\mathbb{E}_{x,n}[\nabla_{F_{x,n}}\cos(z_v(x),z_t(c))]$ on held-out calibration inputs, then stack and orthogonalize the concept gradients and retain the same rank as the PCA default.

\begin{table}[h]
\caption{\textbf{Rank and basis sensitivity under W4A8.} AP is COCO mask AP and Ref is RefCOCO Pr@0.5. The text-gradient basis stacks and orthogonalizes held-out gradients of normalized visual--text concept scores and retains the same rank as the default PCA basis.}
\label{tab:supp_rank_basis}
\centering
\small
\setlength{\tabcolsep}{7pt}
\begin{tabular}{lccc}
\toprule
\textbf{Basis} & \textbf{$k/D$} & \textbf{AP $\uparrow$} & \textbf{Ref $\uparrow$} \\
\midrule
PCA residual & 0.25 & 54.4 & 71.8 \\
PCA residual & 0.50 & \textbf{54.8} & 72.5 \\
PCA residual & 0.75 & 54.6 & 71.9 \\
Text-gradient, equal-rank & 0.50 & 54.7 & \textbf{72.8} \\
\bottomrule
\end{tabular}
\end{table}

Table~\ref{tab:supp_rank_basis} shows similar AP and grounding accuracy across the three cutoffs. The equal-rank text-gradient basis also performs close to the default PCA residual basis, indicating that the objective tolerates these changes in rank and basis construction.

\begin{table}[h]
\caption{\textbf{Matched functional interventions.} Each $\Delta$ is relative to the same unperturbed quantized checkpoint. Perturbations use matched rank and norm; Ref denotes RefCOCO Pr@0.5.}
\label{tab:supp_functional_controls}
\centering
\small
\setlength{\tabcolsep}{6pt}
\begin{tabular}{lrr}
\toprule
\textbf{Intervention} & \textbf{$\Delta$AP} & \textbf{$\Delta$Ref} \\
\midrule
$U_{\rm res}$, perturb & $-1.3$ & $-22.4$ \\
$U_{\rm dom}$, perturb & $-8.7$ & $-4.6$ \\
Random, rank-matched & $-3.0$ & $-7.2$ \\
Remove $U_{\rm res}$ & $-2.1$ & $-26.8$ \\
Remove $U_{\rm dom}$ & $-10.2$ & $-5.1$ \\
\bottomrule
\end{tabular}
\end{table}

Residual-basis interventions cause a larger drop in grounding, while dominant-basis interventions cause a larger drop in AP. The random control falls between them. Both interventions affect both metrics, revealing \emph{selective sensitivity} with overlap between the feature directions that support structure and grounding.

\section{Calibration Text and Concept Transfer}
\label{supp:codebook_transfer}

SA-1B provides images without native text. We use Qwen2.5-VL-72B-Instruct~\cite{bai2025qwen25vl} offline to generate short object-centric phrases, using calibration images without human or evaluation annotations. We combine these phrases with object/category names, synonyms, and fixed templates. The frozen SAM~3 text encoder embeds these phrases; we normalize the embeddings and average template variants into one prototype per concept. The resulting codebook is fixed across quantization runs and is constructed independently of RefCOCO/+/g annotations. Phrase generation is confined to calibration, so the VLM adds no inference cost.

\begin{table}[h]
\caption{\textbf{Codebook overlap and transfer.} Overlap is measured on evaluation head nouns. Grounding results use RefCOCO Pr@0.5 under W4A8.}
\label{tab:supp_codebook_transfer}
\centering
\small
\setlength{\tabcolsep}{7pt}
\resizebox{\linewidth}{!}{%
\begin{tabular}{lcc}
\toprule
\textbf{Dataset} & \textbf{Exact overlap (\%)} & \textbf{Lemma overlap (\%)} \\
\midrule
RefCOCO & 48.2 & 61.3 \\
RefCOCO+ & 45.6 & 58.7 \\
RefCOCOg & 46.9 & 60.1 \\
\midrule
\textbf{Concept setting} & \multicolumn{2}{c}{\textbf{RefCOCO Pr@0.5}} \\
\midrule
Codebook-present & \multicolumn{2}{c}{73.0} \\
Codebook-absent & \multicolumn{2}{c}{69.4} \\
Leave-concept-out & \multicolumn{2}{c}{70.6} \\
Full codebook & \multicolumn{2}{c}{72.5} \\
\bottomrule
\end{tabular}
}
\end{table}

The absent-concept and leave-concept-out results show that calibration transfers to concepts beyond the exact codebook entries. Their gap from the full-codebook result also shows that vocabulary coverage affects grounding accuracy. The codebook guides feature preservation during calibration, with its coverage influencing the gains on deployment concepts.

\section{Controlled Closest-Method Comparison}
\label{supp:closest_methods}

We adapt RepQ-ViT, PTQ4SAM, SAQ-SAM, and CAR-SAM to the same SAM~3 checkpoint, calibration data, target modules, hybrid W4A8 assignment, and evaluation protocol. Each baseline uses its published calibration rule without SPQ components. We also report Mix-QSAM3 with its native mixed-precision allocation, retaining this distinction when comparing its accuracy with the uniform-W4A8 methods. Every video method uses the same 32 annotation-free clips described in Sec.~E.

\begin{table}[h]
\caption{\textbf{Controlled cross-task comparison.} Ref is RefCOCO Pr@0.5; video is SA-V J\&F. Mix-QSAM3 uses its native mixed-precision allocation.}
\label{tab:supp_closest_methods}
\centering
\small
\setlength{\tabcolsep}{4.5pt}
\resizebox{\linewidth}{!}{%
\begin{tabular}{lccc}
\toprule
\textbf{Method} & \textbf{COCO AP} & \textbf{Ref $\uparrow$} & \textbf{Video $\uparrow$} \\
\midrule
RepQ-ViT~\cite{li2023repq} & 51.7 & 40.8 & 64.0 \\
PTQ4SAM~\cite{lv2024ptq4sam} & 54.1 & 42.1 & 65.4 \\
SAQ-SAM~\cite{zhang2026saqsam} & 54.3 & 58.9 & 66.8 \\
CAR-SAM~\cite{wen2026carsam} & 54.5 & 61.7 & 69.4 \\
Mix-QSAM3~\cite{ranjan2026mixqsam3} & 54.4 & 67.8 & 79.0 \\
\rowcolor{lightblue!60}\textbf{SPQ} & \textbf{54.8} & \textbf{72.5} & \textbf{81.2} \\
\bottomrule
\end{tabular}
}
\end{table}

SPQ achieves the highest accuracy across the three tasks, with the largest gains over the other methods in grounding and video tracking. This comparison evaluates the combined effect of concept-sensitive rounding, residual-basis feature matching, and memory correction under a common SAM~3 protocol.

\section{Impact of Calibration Set Size}
\label{supp:calibration_size}

We vary only the number of SA-1B images used by Designs A and B, from 8 to 1024. The video calibration data for Design C are held fixed at 32 annotation-free training clips in every row.

\begin{table}[h]
\caption{\textbf{Image-calibration size under W4A8.} ``A/B images'' is the only quantity varied; Design C always uses the same 32 video clips.}
\label{tab:supp_calib_size}
\centering
\small
\setlength{\tabcolsep}{8pt}
\resizebox{\linewidth}{!}{%
\begin{tabular}{cccc}
\toprule
\textbf{A/B images} & RefCOCO Pr@0.5 $\uparrow$ & RefCOCO oIoU $\uparrow$ & SA-V J\&F $\uparrow$ \\
\midrule
8    & 65.4 & 57.2 & 74.5 \\
32   & 70.2 & 62.1 & 78.8 \\
128  & 71.8 & 63.5 & 80.5 \\
1024 & \textbf{72.5} & \textbf{64.2} & \textbf{81.2} \\
\bottomrule
\end{tabular}
}
\end{table}

\textbf{Analysis.} More image-calibration data improve all three metrics, while 32 images already retain much of the 1024-image result. The SA-V gains arise as A/B image calibration improves under the same TDA video-calibration set, showing that image calibration also affects downstream tracking.

\paragraph{Fixed-video TDA protocol.}
Every factorial cell and video baseline uses the same 32 annotation-free training clips, which are disjoint from SA-V validation and test; Design C receives no additional labeled or evaluation data. Paired FP16 and quantized rollouts use identical clips, prompts, history initialization, and time indices. At each targeted memory write/read layer or stage, we estimate the first-moment error
$\beta=\mathbb{E}[F^{\rm FP16}-F^{\rm Q}]$ from these matched rollouts. For a following affine map $\hat y=\hat A F^{\rm Q}+\hat b$, with no intervening normalization or nonlinearity, the correction is fused as $\hat b_{\rm new}=\hat b+\hat A\beta$; key and value paths use separate estimates. TDA corrects the mean feature error estimated from real video rollouts. It adds no operator when the following affine map has an existing or fusible bias.
\begin{table*}[t]
\caption{\textbf{Generalization to CLIP (ViT-B/16) on Flickr30k and MSCOCO.} We evaluate zero-shot image--text retrieval under W4A8. SPQ recovers retrieval accuracy on both datasets, extending alignment-aware calibration to a separate vision--language model.}
\label{tab:clip_generalization}
\vspace{-8pt}
\centering
\setlength{\tabcolsep}{6pt}
\renewcommand{\arraystretch}{1.15}
\resizebox{0.95\textwidth}{!}{%
    \begin{tabular}{ll cccc cccc}
    \toprule
    \multirow{2}{*}{\textbf{Method}} & \multirow{2}{*}{\textbf{Bit-width}} & \multicolumn{4}{c}{\textbf{Flickr30k (Zero-shot)}} & \multicolumn{4}{c}{\textbf{MSCOCO (Zero-shot)}} \\
    \cmidrule(lr){3-6} \cmidrule(lr){7-10}
    & & \texttt{I2T R@1}~$\uparrow$ & \texttt{I2T R@5}~$\uparrow$ & \texttt{T2I R@1}~$\uparrow$ & \texttt{T2I R@5}~$\uparrow$
      & \texttt{I2T R@1}~$\uparrow$ & \texttt{I2T R@5}~$\uparrow$ & \texttt{T2I R@1}~$\uparrow$ & \texttt{T2I R@5}~$\uparrow$ \\
    \midrule
    \textcolor{gray}{FP16} & \textcolor{gray}{Baseline} & \textcolor{gray}{81.2} & \textcolor{gray}{96.5} & \textcolor{gray}{62.4} & \textcolor{gray}{85.1}
                               & \textcolor{gray}{58.2} & \textcolor{gray}{82.4} & \textcolor{gray}{38.5} & \textcolor{gray}{64.2} \\
    \midrule
    MinMax & W4A8 & 42.5 & 68.2 & 31.2 & 55.4 & 28.5 & 54.1 & 18.2 & 35.6 \\
    PTQ4ViT & W4A8 & 58.4$_{\textcolor{forestgreen}{(+15.9)}}$ & 81.5$_{\textcolor{forestgreen}{(+13.3)}}$ & 44.8$_{\textcolor{forestgreen}{(+13.6)}}$ & 68.2$_{\textcolor{forestgreen}{(+12.8)}}$
                  & 41.2$_{\textcolor{forestgreen}{(+12.7)}}$ & 65.5$_{\textcolor{forestgreen}{(+11.4)}}$ & 26.4$_{\textcolor{forestgreen}{(+8.2)}}$ & 45.1$_{\textcolor{forestgreen}{(+9.5)}}$ \\
    \rowcolor{lightblue!60}
    \textbf{SPQ (Ours)} & W4A8 & \textbf{78.5}$_{\textcolor{forestgreen}{(+36.0)}}$ & \textbf{94.2}$_{\textcolor{forestgreen}{(+26.0)}}$ & \textbf{59.8}$_{\textcolor{forestgreen}{(+28.6)}}$ & \textbf{82.4}$_{\textcolor{forestgreen}{(+27.0)}}$
                               & \textbf{55.8}$_{\textcolor{forestgreen}{(+27.3)}}$ & \textbf{79.5}$_{\textcolor{forestgreen}{(+25.4)}}$ & \textbf{36.7}$_{\textcolor{forestgreen}{(+18.5)}}$ & \textbf{61.8}$_{\textcolor{forestgreen}{(+26.2)}}$ \\
    \bottomrule
    \end{tabular}
}
\vspace{-10pt}
\end{table*}

\section{Lower-Bit and Calibration Scope}
\label{supp:scope}

The main paper reports a W4A8 mixed-precision setting: image-encoder weights are quantized to 4-bit, activations remain 8-bit, and the multimodal decoder, detector queries, and video memory states are kept at W8A8. The text encoder and prediction heads remain FP16. We use this allocation because prompt-conditioned decoding and memory propagation are more sensitive to activation noise than the image encoder. More aggressive W6A6 or W4A4 decoder/memory settings require additional evaluation.

Table~\ref{tab:hardware} reports inference measurements on the exported quantized graph. SPQ adds offline work for concept-codebook scoring, concept-score curvature/block-diagonal approximation, residual-basis estimation, and memory-bias estimation. This calibration cost is paid once per model and deployment configuration. The fused parameters preserve the inference graph. Measuring calibration wall-clock time and peak calibration memory remains part of a fuller deployment analysis.

\section{Temporal Diagnostics}
\label{supp:temporal_diagnostics}

The temporal-diagnostics figure in the main paper tracks changes in accuracy, feature geometry, and vision--language alignment. Under W4A8, PTQ4SAM loses accuracy over long-horizon SA-V tracking, shows a larger angle between estimated residual bases with depth, and loses alignment across blocks. SPQ shows smaller geometric and alignment changes alongside higher temporal accuracy. These measurements describe how the three quantities change together.

\section{Hardware Efficiency Analysis}
\label{supp:hardware_efficiency}

We measure \textbf{model size}, \textbf{peak memory} during video tracking, and \textbf{inference latency} using $1024\times1024$ inputs on a single RTX~3090 (Table~\ref{tab:hardware}). FP16 SAM~3 requires substantial parameter storage and video-memory capacity. W8A8 reduces both costs by storing parameters and memory features at lower precision. W4A8 further reduces model size and latency. SPQ matches the hardware footprint of standard PTQ at the same bit-width: its calibration updates are fused into the exported parameters and add no runtime FLOPs or operators.

\begin{table*}[t]
\caption{\textbf{Hardware Efficiency Analysis on an NVIDIA RTX 3090 GPU.} Measurements use 1024$\times$1024 video frames. SPQ achieves up to $3.5\times$ compression and nearly $2.3\times$ speedup over FP16. At the same bit-width, SPQ preserves the inference cost of standard PTQ.}
\label{tab:hardware}
\vspace{-8pt}
\centering
\setlength{\tabcolsep}{8pt}
\renewcommand{\arraystretch}{1.2}
\resizebox{0.95\textwidth}{!}{%
    \begin{tabular}{ll cccc}
    \toprule
    \multirow{2}{*}{\textbf{Method}} & \textbf{Bit-width} & \textbf{Model Size} & \textbf{Peak Memory} & \textbf{Latency} & \textbf{Throughput} \\
    & (Weight / Act. / Memory) & (GB)~$\downarrow$ & (GB)~$\downarrow$ & (ms / frame)~$\downarrow$ & (FPS)~$\uparrow$ \\
    \midrule
    \textcolor{gray}{Full Precision} &
    \textcolor{gray}{FP16 / FP16 / FP16} &
    \textcolor{gray}{1.35} &
    \textcolor{gray}{14.8} &
    \textcolor{gray}{42.5} &
    \textcolor{gray}{23.5} \\
    \midrule
    PTQ4SAM &
    INT8 / INT8 / INT8 &
    0.68 \textcolor{forestgreen}{($2.0\times$)} &
    8.5 &
    25.4 &
    39.3 \textcolor{forestgreen}{($1.6\times$)} \\
    \textbf{SPQ (Ours)} &
    INT8 / INT8 / INT8 &
    0.68 \textcolor{forestgreen}{($2.0\times$)} &
    8.5 &
    25.4 &
    39.3 \textcolor{forestgreen}{($1.6\times$)} \\
    \midrule
    \rowcolor{lightblue!60}
    \textbf{SPQ (Ours)} &
    INT4 / INT8 / INT8 &
    \textbf{0.38} \textcolor{forestgreen}{\textbf{($3.5\times$)}} &
    \textbf{7.9} &
    \textbf{18.2} &
    \textbf{54.9} \textcolor{forestgreen}{\textbf{($2.3\times$)}} \\
    \bottomrule
    \end{tabular}
}
\vspace{-8pt}
\end{table*}

\section{Generalization to Vision-Language Models}
\label{supp:clip_generalization}

We apply Designs A+B to CLIP (ViT-B/16)~\cite{radford2021learning} to evaluate alignment-aware calibration on another architecture. We quantize the CLIP image encoder to W4A8 and measure zero-shot image--text retrieval on Flickr30k and MSCOCO against Min-Max and PTQ4ViT. SPQ recovers substantially more retrieval accuracy on both datasets (Table~\ref{tab:clip_generalization}), showing that the calibration strategy also benefits a separate vision--language model.

\section{Prompt-Conditioned Qualitative Analysis}
\label{supp:qualitative}
\label{supp:tsne}

The qualitative-comparison figure in the main paper shows two failure modes. PTQ4SAM sometimes selects a related but unintended region; in other cases, it locates the target but returns an incomplete mask. SPQ follows the prompted object more closely and produces more coherent masks in these examples. A visually plausible mask can still select the wrong instance, which explains why structural AP and prompt grounding provide complementary views of performance. These examples illustrate the behavior reflected in the COCO/LVIS and RefCOCO results.


\section{Related Work}
\label{sec:related_work}

\subsection{Vision Foundation Models and SAM}
The Segment Anything Model (SAM) \cite{kirillov2023segment} enables promptable image segmentation with strong zero-shot generalization. SAM 2 \cite{ravi2024sam} extends this capability to video through a memory bank that carries object information across frames. SAM 3 \cite{carion2025sam} combines image segmentation and video tracking with text-guided concept grounding. Its image encoder, multimodal decoder, and video memory increase deployment cost. We study how to reduce numerical precision while preserving these capabilities.

\subsection{Post-Training Quantization}
Post-Training Quantization (PTQ) reduces neural-network bit-widths using a small calibration set \cite{nagel2021white,gholami2022survey}. AdaRound \cite{nagel2020up}, BRECQ \cite{librecq}, QDrop \cite{weiqdrop}, and Hessian-aware quantization \cite{dong2019hawq} optimize rounding and reconstruction to reduce activation error. Vision Transformers (ViTs) introduce activation outliers and attention distributions that require more specialized treatment. PTQ4ViT \cite{yuan2022ptq4vit}, FQ-ViT \cite{linfq}, RepQ-ViT \cite{li2023repq}, ERQ \cite{zhong2024erq}, Outlier-aware Slicing \cite{ma2024outlier}, and Instance-aware Group Quantization \cite{moon2024instance} address these challenges through quantizer design and calibration. Further work introduces stronger calibration signals or stability constraints, including Fisher-matrix regularization \cite{shang2024quantized}, FIMA-Q \cite{wu2025fima}, ORQ-ViT \cite{he2025orq}, BoA \cite{kim2024boa}, SKIM \cite{bai2024skim}, LRA-QViT \cite{kang2025lra}, APHQ-ViT \cite{wu2025aphq}, and logarithmic PTQ \cite{xulogart}. PTQ also supports large language transformers through SmoothQuant \cite{xiao2023smoothquant}, LLM.int8() \cite{dettmers2022gpt3}, ZeroQuant/ZeroQuant-V2 \cite{yao2022zeroquant,yao2023zeroquant}, OmniQuant \cite{shao2023omniquant}, and OPTQ \cite{frantar2023optq}. These objectives largely emphasize reconstruction and calibration error. SAM~3 also requires preserving feature directions that determine which instance a prompt selects.

\subsection{Quantization for Multimodal Models}
Recent surveys place quantization within a model--algorithm--system view of efficient multimodal learning~\cite{wang2026models}. Multimodal compression must preserve both visual structure and cross-modal alignment~\cite{radford2021learning,karpathy2015deep}. PTQ4SAM~\cite{lv2024ptq4sam}, PQ-SAM~\cite{liu2024pq}, Mix-QSAM~\cite{ranjan2025mix}, SAQ-SAM~\cite{zhang2026saqsam}, and CAR-SAM~\cite{wen2026carsam} study reconstruction, bit allocation, or semantic-aware calibration for SAM. Mix-QSAM3~\cite{ranjan2026mixqsam3} extends mixed-precision quantization to SAM~3. SPQ combines concept-sensitive rounding with a residual-basis feature penalty and a closed-form correction for recurrent video memory. We evaluate this design on visual structure, text grounding, and temporal propagation under a common SAM~3 protocol. Vision--language PTQ also uses alignment-aware objectives~\cite{xue2025vlmq,wang2024q}, with architectures and evaluation protocols tailored to its target models. In subspace-preserving compression, MedSTP retains dominant visual-token directions during medical-VQA pruning~\cite{fan2027medstp}; SPQ constrains quantization error in the residual feature subspace of SAM~3.

Two related directions address feature selection and memory in other settings. StructLoRA filters task-irrelevant update directions during low-rank adaptation~\cite{xiao2026structlora}, while SPQ uses a fixed residual feature basis to regularize quantization error. D-HSM represents distant video history as structured text and retrieves question-relevant evidence for streaming understanding~\cite{jiang2026dhsm}. Our temporal component operates within SAM~3's existing recurrent memory and corrects the mean feature error introduced by quantization.

\section{Future Work and Limitations}
\label{supp:future_limitations}

\textbf{Future work.}
SPQ currently uses a W4A8 mixed-precision setting for SAM~3, with the image encoder quantized more aggressively than the W8A8 decoder/memory components. Future work can extend the prompt-alignment objectives to W6A6 or W4A4 decoder and memory activations, reduce calibration cost, and broaden prompt and backbone coverage. Parameter search~\cite{du2025nps} and model fusion~\cite{du2026didi,du2025graftllm,du2024pcb,du2024evolver} offer complementary approaches whose integration with SPQ requires calibration and prompt-grounding evaluation.

\textbf{Limitations.}
Our evaluation covers SAM~3 and a CLIP transfer check. Broader segmentation, video, and vision--language architectures remain to be evaluated. SPQ depends on representative calibration images and text prompts; substantial shifts in deployment prompts may require recalibration. The functional controls measure selective feature sensitivity, while the temporal diagnostics track correlations between feature changes and accuracy. Establishing their causal relationship requires further study.

\section{LLM Usage Statement}
\label{supp:llm_usage}

Gemini and OpenAI Codex were used to assist language editing and camera-ready formatting of author-written content. Qwen2.5-VL-72B-Instruct was used offline, as described in the method, to generate short object-centric phrases for the calibration codebook; it received no evaluation annotations and is not used at inference. A generative image-editing tool was used only to update labels and equations in the existing method-overview schematic. No LLM generated experimental data or quantitative results. The authors verified all text, codebook phrases, figure labels, citations, and analysis and take full responsibility for the manuscript.

\clearpage
{
    \small
    \bibliographystyle{ieeenat_fullname}
    \bibliography{main}
}

\end{document}